%% file: paper.tex
\def\year{2021}\relax
\documentclass[letterpaper]{article} 
\usepackage{aaai21}  
\usepackage{times}  
\usepackage{helvet} 
\usepackage{courier}  
\usepackage[hyphens]{url}  
\usepackage{graphicx} 
\usepackage{natbib}  
\usepackage{caption} 
\nocopyright

\usepackage{xspace}
\usepackage[center]{subfigure} 
\usepackage{tabularx}
\usepackage{tablefootnote}
\usepackage{threeparttable}
\usepackage{multirow}
\usepackage{array}
\usepackage{booktabs} 
\usepackage{makecell}
\usepackage{enumitem}
\usepackage{amsmath}
\usepackage{amssymb}
\usepackage[ruled,linesnumbered]{algorithm2e} 
\usepackage[]{algorithmic} 
\usepackage{balance}
\usepackage{xcolor}
\usepackage{mathrsfs}
\usepackage{mathtools}
\usepackage{siunitx}

\usepackage[title, page]{appendix}

\begin{document}

\input{dfn}
%
\title{\method: Learning to Walk across Time for Temporal Knowledge Graph Completion}
\author{Jaehun Jung\textsuperscript{1},\hspace{1mm}
Jinhong Jung\textsuperscript{2},\hspace{1mm}
U Kang\textsuperscript{1}\\\vspace{2mm}\hspace{1mm}
{\normalfont\normalsize \textsuperscript{1}{Department of Computer Science, Seoul National University}\\\vspace{1mm}
\textsuperscript{2}{Department of Computer Science, Jeonbuk National University}\\\vspace{1mm}
sharkmir1@snu.ac.kr,
jinhongjung@jbnu.ac.kr, 
ukang@snu.ac.kr}}
\maketitle

\begin{abstract}
    \input{000abstract}
\end{abstract}

\section{Introduction}
\label{sec:intro}
\input{010intro}

\section{Related Work}
\label{sec:related}
\input{020related}

\section{Proposed Method}
\label{sec:method}
\input{030method}

\section{Experiment}
\label{sec:experiment}
\input{040experiment}

\section{Conclusion}
\label{sec:conclusion}
\input{050conclusion}

\bibliography{BIB/myref}
\input{060appendix}

\end{document}

%% file: dfn.tex
\newtheorem{observation}{Observation}
\newtheorem{problem}{Problem}
\newtheorem{definition}{Definition}
\newtheorem{lemma}{Lemma}
\newtheorem{theorem}{Theorem}
\newtheorem{proof}{Proof}

\renewcommand{\algorithmicrequire}{\textbf{Input:}}
\renewcommand{\algorithmicensure}{\textbf{Output:}}

\newcommand*{\QEDA}{\hfill\ensuremath{\blacksquare}}%
\newcommand*{\QEDB}{\hfill\ensuremath{\Box}}%

\newcommand{\mat}[1]{\mathbf{#1}}
\newcommand{\matt}[1]{\mathbf{#1}^{\top}}
\newcommand{\mattt}[1]{\mathbf{\tilde{#1}}^{\top}}
\newcommand{\mati}[1]{\mathbf{#1}^{-1}}
\newcommand{\vect}[1]{\mathbf{#1}}
\newcommand{\vectt}[1]{\mathbf{#1}^{\top}}
\newcommand{\gvect}[1]{\mathbf{\boldsymbol#1}}
\newcommand{\set}[1]{#1}

\newcommand{\method}[0]{\textsc{T-GAP}\xspace}
\newcommand{\methodlong}[0]{\text{Temporal GNN with Attention Propagation}\xspace}

\newcommand{\ON}[1]{\small\overrightarrow{\set{N}}_{#1}\normalsize}
\newcommand{\IN}[1]{\small\overleftarrow{\set{N}}_{#1}\normalsize}

\newcommand{\citesupp}{(Supplementary material)}

\newcommand{\T}[0]{\mathcal{T}} 

%% file: 000abstract.tex
Temporal knowledge graphs (TKGs) inherently reflect the transient nature of real-world knowledge, as opposed to static knowledge graphs. 
Naturally, automatic TKG completion has drawn much research interests for a more realistic modeling of relational reasoning. 
However, most of the existing models for TKG completion extend static KG embeddings that do not fully exploit TKG structure, thus lacking in 1) accounting for temporally relevant events already residing in the local neighborhood of a query, and 2) path-based inference that facilitates multi-hop reasoning and better interpretability. 
In this paper, we propose \method, a novel model for TKG completion that maximally utilizes both temporal information and graph structure in its encoder and decoder. 
\method encodes query-specific substructure of TKG by focusing on the temporal displacement between each event and the query timestamp, and performs path-based inference by propagating attention through the graph. 
Our empirical experiments demonstrate that \method not only achieves superior performance against state-of-the-art baselines, but also competently generalizes to queries with unseen timestamps. 
Through extensive qualitative analyses, we also show that \method enjoys from transparent interpretability, and follows human intuition in its reasoning process.

%% file: 010intro.tex
Knowledge graph (KG), due to its expressiveness over structured knowledge, has been widely used in various applications including recommender system \cite{explainablerec}, information retrieval \cite{ednr}, synonym discovery \cite{KangPHF12,journals/debu/PapalexakisKFSH13}, concept discovery \cite{conf/icde/JeonPKF15,journals/vldb/JeonPFSK16}, and question answering \cite{variationalqa}. Moreover, the inherent sparseness of KGs gave rise to research interests over automatic knowledge graph completion, predicting missing entity for incomplete queries in form of (subject, predicate, ?).


Recent advancements in KG completion tasks have extended to a more challenging domain of temporal knowledge graphs (TKGs), as they model realistic events that are only temporarily valid. Triples in temporal graphs are annotated with corresponding time token, taking form of (subject, predicate, object, timestamp). Naturally, TKG completion task can be formulated as predicting missing tail entity for queries in form of (subject, predicate, ?, timestamp).

Majority of existing approaches to TKG completion propose a straightforward extension of conventional KG embeddings onto the temporal graphs \cite{hyte,tcomplex}. Although these approaches are generally successful in TKG completion, we observe two major rooms of improvement from the existing models, each from the encoding phase, and the decoding phase.

\begin{figure}
  \centering
  \includegraphics[width=.4\textwidth]{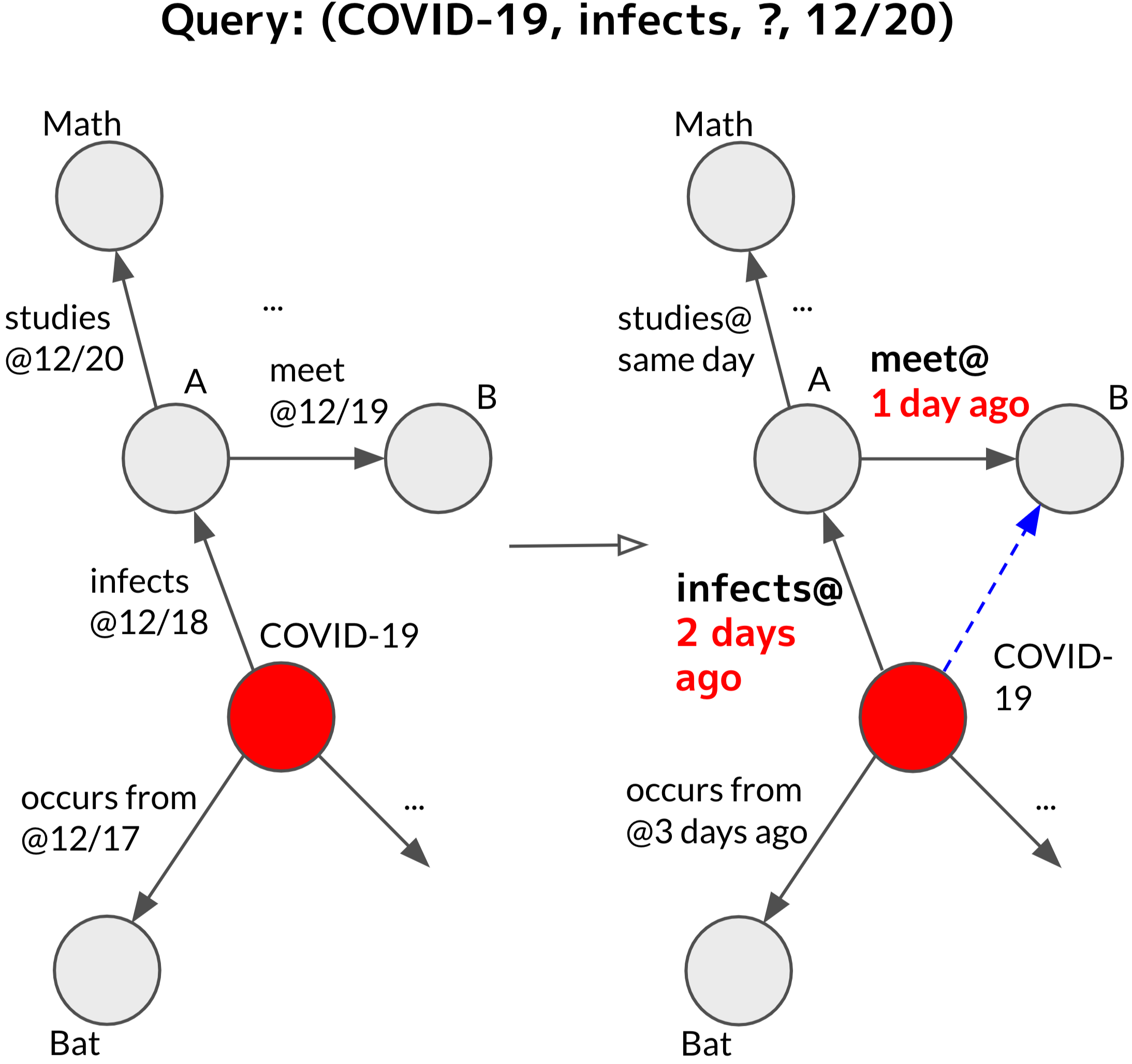}
  \caption{Example of temporal displacement. Edges in bold are important events relevant to the input query.}
  \label{fig:motivation}
  \vspace{-3mm}
\end{figure}

In the encoding phase, a model could benefit from the rich neighborhood information residing in the structure of TKGs. Extracting, and encoding query-relevant information from the neighborhood nodes and their associated edges would help in fine-grained modeling of entity representation. The importance of neighborhood encoding has already been appreciated in static KGs \cite{sacn,kbgat}, but extension of these models to TKG is non-trivial, due to the additional time dimension in each triple.

Next, in the decoding phase, relational reasoning on TKG could leverage path-based inference. Several works adopted path-traversal model in static KGs \cite{dpmpn,minerva}, showing preferable performance in relational reasoning compared to embedding-based models. Although path-based inference helps in capturing long-term dependency between nodes and gives better interpretability over model's reasoning process, these approaches are yet to be examined in TKG completion tasks.

To this end, we propose \method (\textbf{T}emporal \textbf{G}NN with \textbf{A}ttention \textbf{P}ropagation), a novel model for TKG completion, that tackles both challenges stated above. In the encoder, we introduce a new type of temporal graph neural network (GNN), which attentively aggregates query-relevant information from each entity’s local neighborhood.
Specifically, we focus on encoding the temporal displacement between the timestamps of the input query and each edge being encoded. An intuitive example is presented in Figure 1. Evidently, the two most important facts to answer the given query (COVID-19, infects, ?, 12/20), are that A has been infected to COVID-19 at 12/18, and A met B at 12/19. Here, one should note that the valuable information lies in the fact that A got infected \textit{2 days before} the time of interest, not that he was infected at a \textit{specific day} of 12/18. What matters most when accounting for temporal events, is the relative displacement between the event and the time of interest, rather than the absolute time of the event. To effectively capture the temporal displacement, our proposed encoder separately encodes both the sign of the displacement (i.e. whether the time of the event belongs to past, present, or future), and the magnitude of the displacement (i.e. how far is the event from the time of our interest).

Also, \method performs a generalized path-based inference over TKG, based on the notion of Attention Flow \cite{attnFlow}. In each decoding step, our model explores KG by propagating attention value at each node to its reachable neighbor nodes, rather than sampling one node to walk from the neighborhood. The soft approximation of path traversal with attention propagation not only allows our model to be easily trained with end-to-end supervised learning, but also provides better interpretation over its reasoning process, compared to embedding-based models.

In summary, our contributions are as follows:
\begin{itemize}
    \item We introduce a new GNN encoder that effectively captures query-relevant information from temporal KGs.
    \item Based on the encoder, we present \method, a novel path-based TKG reasoning model. We examine \method in 3 benchmark datasets in TKG completion task, and the quantitative metrics show clear improvement in all benchmarks compared to the state-of-the-art baselines.
    \item By analyzing the inferred attention distribution, we show that \method possesses clear interpretability over its reasoning process, which has not yet been extensively discussed in TKG domain.
\end{itemize}

%% file: 020related.tex
Various approaches have been made toward automatic completion of static KGs. Majority of conventional approaches propose embedding-based models, including translative \cite{transE,transH}, and factorization-based models \cite{distmult,complex}. To complement for the weak representation power of KG embeddings, several recent works incorporate neural network either to the scoring function, or as an additional encoding layer. ConvE \cite{convE} adopts convolutional layer to model sophisticated interaction between entities in the scoring function. KBGAT \cite{kbgat}, and RGHAT \cite{rghat} adopt a variant of graph attention network to contextualize entity embedding with the corresponding neighborhood structure. DPMPN \cite{dpmpn} employs two GNNs to encode both the original graph and an induced subgraph, for a scalable learning of KG structure. We extend the neighborhood encoding scheme of these prior works to temporal graphs, especially focusing on query dependent encoding of KG structure.

Existing works on TKG completion primarily focus on extending static KG embedding to dynamic graphs. Different models mainly differ in how to represent independent timestamps, and incorporate it to their scoring functions. HyTE \cite{hyte} extends TransH \cite{transH}, projecting entity and relation embedding to time-specific hyperplane. Garc{\'\i}a-Dur{\'a}n et al. (\citeyear{ta-distmult}) propose to represent temporal relation as a sequence of relation type and characters in the timestamp, and encode the sequence using RNN. TComplEx \cite{tcomplex} considers the score of each triple as canonical decomposition of order 4 tensors in complex domain, adding time embedding to the order 3 decomposition of ComplEx.  \citet{de-simple} suggest to learn entity representation that changes over time, transforming part of the embedding with sinusoidal activation of learned frequencies.

Meanwhile, path-based reasoning has been actively employed for node prediction on knowledge graphs. \citet{PtransE} infuse multi-hop path information into entity representation, using additive composition of relation embeddings. \citet{minerva} and \citet{kgqa} consider KG reasoning problem as partially observed Markov Decision Process, training a policy network that starts traversing from the query head and reaches at the predicted tail entity. \citet{attnFlow} propose a soft approximation of path traversal with attention distribution. \method aligns with these line of works by exploring informative paths relevant to the input query with attention propagation, to maximally utilize the KG structure during the decoding process.

Lastly, we find connections to our work from recently suggested GNNs for dynamic graphs. \citet{evolvegcn} propose a variant of graph convolutional network (GCN) suited for TKG, employing RNN to evolve GCN parameters across time. \citet{hawkes} present Graph Hawkes Neural Network, modeling temporal dependency between events with Hawkes process. While these works focus on modeling the evolution of the graph as whole, we discuss a new methodology of encoding temporal displacement between each event and the input query, which better suits our goal to explore query-relevant paths and reach the answer node.

%% file: 030method.tex
\lineskip=0pt 

\begin{figure*}[ht]
  \centering
  \includegraphics[width=\textwidth]{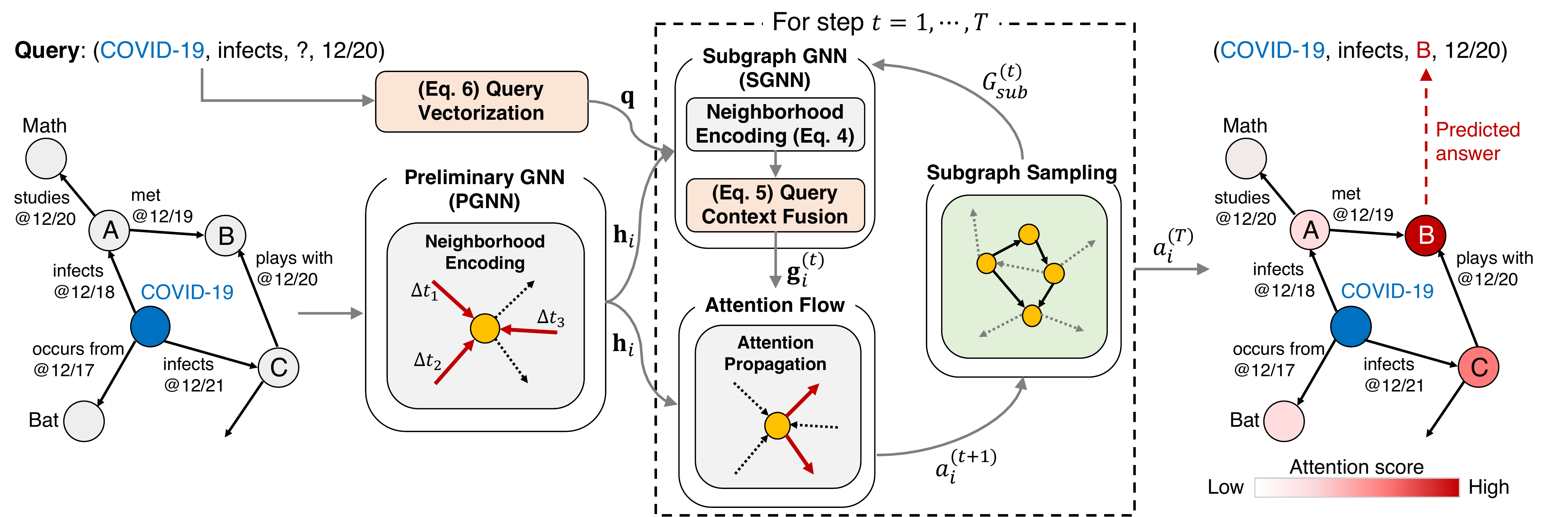}
  \caption{Overview of \method. Starting from the query head, \method explores relevant nodes and edges by iteratively propagating attention, and reaches at the target entity after the final propagation step.}
  \label{fig:architecture}
\end{figure*}

\subsection{Overview}
First, we denote TKG as $G_{KG} = \{(v, r, u, t)\} \subseteq V_{KG} \times \set{R}_{KG} \times \set{V}_{KG} \times \set{T}_{KG}$, where $\set{V}_{KG}$ is a set of entities, $\set{R}_{KG}$ is a set of relations, and $\set{T}_{KG}$ is a set of timestamps associated with the relations. Given the graph $G_{KG}$ and query $\vect{q} = (v_{query}, r_{query}, ?, t_{query})$, TKG completion is formulated as predicting $u \in \set{V}_{KG}$ most probable to fill in the query. We also denote $\IN{i}$ as a set of incoming neighbor nodes of $v_i$, i.e. nodes posessing edges toward $v_i$, and $\ON{i}$ as a set of outgoing neighbor nodes of $v_i$.

Figure \ref{fig:architecture} illustrates an overview of \method's reasoning process. \method consists of 4 sub-modules: Preliminary GNN (PGNN), Subgraph GNN (SGNN), Subgraph Sampling, and Attention Flow. Given $G_{KG}$ and the query, in the encoding phase, \method first uses PGNN to create preliminary node feature $\vect{h}_i$ for all entities in the $G_{KG}$.

Next, at each decoding step $t = 1, \cdots, T$, \method iteratively samples a subgraph $G^{(t)}_{sub}$ from $G_{KG}$, that consists only of query-relevant nodes and edges.
For each entity $i$ included in $G^{(t)}_{sub}$, SGNN creates query-dependent node feature $\vect{g}_i^{(t)}$, incorporating the query vector $\vect{q}$ and the preliminary feature $\vect{h}_i$.
As noted in \citet{dpmpn}, this additional encoding of induced subgraph not only helps in extracting only the query-relevant information from the original graph, but also scales up GNN by input-dependent pruning of irrelevant edges.
Using both $\vect{h}_i$ and $\vect{g}_i^{(t)}$, Attention Flow computes transition probability to propagate the attention value of each node to its reachable neighbor nodes, creating the next step's node attention distribution $a_i^{(t+1)}$. After the final propagation step $T$, the answer to the input query is inferred as the node with the highest attention value $a^{(T)}_i$.

\subsection{Preliminary GNN}
Given $G_{KG}$, \method first randomly initializes node feature $\vect{h}_i$ for all $v_i \in V_{KG}$. Then, to contextualize the representation of entities in $G_{KG}$ with the graph structure, each layer in PGNN updates node feature $\vect{h}_i$ of entity $v_i$ by attentively aggregating $v_i$'s neighborhood information.
The important intuition underlying PGNN is that the temporal displacement between timestamps of the query and each event is integral to capture the time-related dynamics of each entity.
Therefore, for each timestamp $t_e$ of edge $e$ in $G_{KG}$, we resolve to separately encode the sign and magnitude of the temporal displacement $\triangle t_{e} = t_e - t_{query}$.
Concretely, PGNN computes message $\vect{m}_{ij}$ from entity $v_i$ to $v_j$ as follows:
\begin{equation} \label{eq:1}
    \begin{split}
    &\vect{m}_{ij} = \mat{W}_{\lambda(\triangle t_{ij})}(\vect{h}_i + \gvect{\rho}_{ij} + \gvect{\tau}_{|\triangle t_{ij}|}) \\
    &\text{where}  \,\, \mat{W}_{\lambda(\triangle t_{ij})} =
    \begin{cases}
        \mat{W}_{past} & \text{if } \triangle t_{ij} < 0\\
        \mat{W}_{present} & \text{if } \triangle t_{ij} = 0\\
        \mat{W}_{future} & \text{if } \triangle t_{ij} > 0
    \end{cases}
    \end{split}
\end{equation}
$\gvect{\rho}_{ij}$ is a relation-specific parameter associated with $r_{ij}$ which denotes the relation that connects $v_i$ to $v_j$.
In addition to the entity and relation, we learn the discretized embedding of size of temporal displacement, i.e. $\gvect{\tau}_{|\triangle t_{ij}|}$.
We take account of the sign of displacement by applying sign-specific weight for each event.

Next, the new node feature $\vect{h}'_j$ is computed by attention weighted sum of all incoming messages to $v_j$:
\begin{equation} \label{eq:2}
    \begin{split}
    &\vect{h}'_j = \sum_{i \in \IN{j}} a_{ij} \vect{m}_{ij}, \\
    &a_{ij} = \texttt{softmax}_i(\alpha_{ij}),\\
    &\alpha_{ij} = \texttt{LeakyReLU}\left((\mat{W}_Q \vect{h}_j)^{\top} (\mat{W}_K \vect{m}_{ij})\right)
    \end{split}
\end{equation}
The attention values are computed by applying softmax over all incoming edges of $v_j$, with $\vect{h}_j$ as query and $\vect{m}_{ij}$ as key.

In addition, we extend this attentive aggregation scheme to multi-headed attention, which helps to stabilize the learning process and jointly attend to different representation subspaces~\cite{gat}. Hence our message aggregation scheme is modified to:
\begin{equation} \label{eq:3}
    \vect{h}'_j = \overset{K}{\underset{k=1}{\big{\|}}} \sum_{i \in \IN{j}} a^{k}_{ij} \vect{m}^{k}_{ij}
\end{equation}
concatenating independently aggregated neighborhood features from each attention heads, where $K$ is a hyperparameter indicating the number of attention heads.

\subsection{Subgraph GNN}
At each decoding step $t$, SGNN updates node feature $\vect{g}_i$ for all entities that are included in the induced subgraph of current step, $G^{(t)}_{sub}$.
We present the detailed procedure of subgraph sampling in upcoming section.
Essentially, SGNN not only contextualizes $\vect{g}_i$ with respective incoming edges, but also infuses the query context vector with the entity representation.
First, the subgraph features for entities newly added to the subgraph, are initialized to their corresponding preliminary features $\vect{h}_j$.
Next, SGNN performs message propagation, using the same message computation and aggregation scheme as PGNN (Eq. \ref{eq:1}-\ref{eq:3}), but with separate parameters:
\begin{equation}
    \vect{\widetilde{g}}'_j = \overset{K}{\underset{k=1}{\big{\|}}} \sum_{i \in \IN{j}} a^{k}_{ij} \vect{m}^{k}_{ij}
\end{equation}
This creates an intermediate node feature $\vect{\widetilde{g}}'_j$.
The intermediate features are then concatenated with query context vector $\vect{q}$, and linear-transformed back to the node embedding dimension, creating new feature $\vect{g}'_j$:
\begin{equation}
    \vect{g}'_j = \mat{W}_g[\vect{\widetilde{g}}'_j\, \| \,\vect{q}]
\end{equation}
\begin{equation}
    \vect{q} \! = \! \mat{W}_c \! \times \! \texttt{LeakyReLU}\!\left(\mat{W}_{present}(\vect{h}_{query} \! + \! \gvect{\rho}_{query})\right)
\end{equation}
where $\vect{h}_{query}$ is the preliminary feature of $v_{query}$, and $\gvect{\rho}_{query}$ is the relation parameter for $r_{query}$.


\subsection{Attention Flow}
\method models path traversal with the soft approximation of attention flow, iteratively propagating the attention value of each node to its outgoing neighbor nodes. Initially, the node attention is initialized to 1 for $v_{query}$, and 0 for all other entities. Hereafter, at each step $t$, Attention Flow propagates edge attention $\widetilde{a}^{(t)}_{ij}$ and aggregates them to node attention $a^{(t)}_j$:
\begin{equation*}
    \begin{split}
        &\widetilde{a}^{(t+1)}_{ij} = \T^{(t+1)}_{ij} a^{(t)}_i, \,\, a^{(t+1)}_j = \sum_{i \in \IN{j}}\widetilde{a}^{(t+1)}_{ij}\\
        &s.t. \,\, \sum_i a^{(t+1)}_i = 1, \,\, \sum_{ij} \widetilde{a}^{(t+1)}_{ij} = 1
    \end{split}
\end{equation*}
The key here is the transition probability $\T_{ij}$. In this work, we define $\T_{ij}$ as applying softmax over the sum of two scoring terms, regarding both the preliminary feature $\vect{h}$, and the subgraph feature $\vect{g}$:
\begin{equation*}
    \begin{split}
        \T_{ij}^{(t+1)} = \texttt{softmax}_j(&score(\vect{g}^{(t)}_i, \vect{g}^{(t)}_j, \gvect{\rho}_{ij}, \gvect{\tau}_{|\triangle t_{ij}|})\, +\\
                  &score(\vect{g}^{(t)}_i, \vect{h}_j, \gvect{\rho}_{ij}, \gvect{\tau}_{|\triangle t_{ij}|})),\\
    \end{split}
\end{equation*}
\begin{equation*}
        score(\vect{i}, \vect{j}, \vect{r}, \gvect{\tau}) = \sigma\left((\mat{W}_Q \vect{i})^{\top}(\vect{W}_K (\vect{j} + \vect{r} + \gvect{\tau}))\right)
\end{equation*}
The first scoring term accounts only for subgraph feature $\vect{g}_i$ and $\vect{g}_j$, giving additional point to entities that are already included in the subgraph
(note that $\vect{g}_i$ is initialized to zero for entities not yet included in the subgraph).
Meanwhile, the second scoring term could be regarded as \textit{exploring term}, as it relatively prefers entities not included in the subgraph, by modeling the interaction between $\vect{g}_i$ and $\vect{h}_j$.

As \method consists only of differentiable operations, the path traversal of \method can be trained end-to-end by directly supervising on the node attention distribution after $T$ propagation steps.
We train \method to maximize the log probability of the answer entity $u_{label}$ at step $T$.
\begin{equation*}
    \mathcal{L} = -\log{a^{(T)}_{u_{label}}}
\end{equation*}

\input{table/performance}

\subsection{Subgraph Sampling}
The decoding process of \method depends on the iterative sampling of query-relevant subgraph $G^{(t)}_{sub}$. The initial subgraph $G^{(0)}_{sub}$ before the first propagation step contains only one node, $v_{query}$. As the propagation step proceeds, edges with high relevance to the input query, measured by the size of attention value assigned to the edges, are added to the previous step's subgraph.
Specifically, the subgraph sampling at step $t$ proceeds as follows:

\begin{itemize}
    \item Find $x$ number of \textit{core nodes}  with highest (nonzero) node attention value $a^{(t-1)}_i$ at the previous step.
    \item For each of the core node, sample $y$ number of edges that originate from the node.
    \item Among $x \cdot y$ sampled edges, find $z$ number of edges with highest edge attention value $\widetilde{a}^{(t)}_{ij}$ at the current step.
    \item Add the $z$ edges to $G^{(t-1)}_{sub}$.
\end{itemize}
In this module, $x, y, z$ are hyperparameters.
Intuitively, we only collect `important' events that originate from `important' entities (core nodes) with respect to the query, while keeping the subgraph size under control (edge sampling). We provide an illustrative example on subgraph sampling in Appendix A.

Note that although edge sampling brings in stochasticity to \method's inference, this does not hinder the end-to-end training of the model. Since the sampling is not parameterized and we only use node feature $\vect{g}$ from the sampled subgraph, gradients back-propagate through $\vect{g}$, not through the sampling operation.


%% file: table/performance.tex
\begin{table*}[t]
    \resizebox{\textwidth}{!}{\begin{tabular}{ cccccccccccccc }
        \Xhline{2\arrayrulewidth}
        \multicolumn{2}{c|}{} & \multicolumn{4}{c}{\textbf{ICEWS14}} & \multicolumn{4}{c}{\textbf{ICEWS05-15}} & \multicolumn{4}{c}{\textbf{Wikidata11k}} \\
        \multicolumn{2}{c|}{Model} & \textit{MRR} & \textit{Hits@1} & \textit{Hits@3} & \multicolumn{1}{c|}{\textit{Hits@10}} & \textit{MRR} & \textit{Hits@1} & \textit{Hits@3} & \multicolumn{1}{c|}{\textit{Hits@10}} & \textit{MRR} & \textit{Hits@1} & \textit{Hits@3} & \textit{Hits@10} \\
        \hline\hline
        \multicolumn{2}{c|}{TransE [$\blacktriangledown$]} & 0.280 & 9.4 & - & \multicolumn{1}{c|}{63.7}
        & 0.294 & 9.0 & - & \multicolumn{1}{c|}{66.3} & 0.316 & 18.1 & - & 65.9 \\
        \multicolumn{2}{c|}{DistMult [$\blacktriangledown$]} & 0.439 & 32.3 & - & \multicolumn{1}{c|}{67.2}
        & 0.456 & 33.7 & - & \multicolumn{1}{c|}{69.1} & 0.316 & 18.1 & - & 66.1 \\
        \hline
        \multicolumn{2}{c|}{ConT [$\diamond$]} & 0.185 & 11.7 & 20.5 & \multicolumn{1}{c|}{31.5}
        & 0.163 & 10.5 & 18.9 & \multicolumn{1}{c|}{27.2} & - & - & - & - \\
        \multicolumn{2}{c|}{TTransE [$\blacktriangledown$]} & 0.255 & 7.4 & - & \multicolumn{1}{c|}{60.1}
        & 0.271 & 8.4 & - & \multicolumn{1}{c|}{61.6} & 0.488 & 33.9 & - & 80.6 \\
        \multicolumn{2}{c|}{HyTE [$\diamond$]} & 0.297 & 10.8 & 41.6 & \multicolumn{1}{c|}{65.5}
        & 0.316 & 11.6 & 44.5 & \multicolumn{1}{c|}{68.1} & - & - & - & - \\
        \multicolumn{2}{c|}{TA-TransE [$\blacktriangledown$]} & 0.275 & 9.5 & - & \multicolumn{1}{c|}{62.5}
        & 0.299 & 9.6 & - & \multicolumn{1}{c|}{66.8} & 0.484 & 32.9 & - & 80.7 \\
        \multicolumn{2}{c|}{TA-DistMult [$\blacktriangledown$]} & 0.477 & 36.3 & - & \multicolumn{1}{c|}{68.6}
        & 0.474 & 34.6 & - & \multicolumn{1}{c|}{72.8} & 0.700 & 65.2 & - & 78.5 \\
        \multicolumn{2}{c|}{DE-SimplE [$\diamond$]} & 0.526 & 41.8 & 59.2 & \multicolumn{1}{c|}{72.5}
        & 0.513 & 39.2 & 57.8 & \multicolumn{1}{c|}{74.8} & 0.310 & 18.4 & 31.8 & 62.5 \\
        \multicolumn{2}{c|}{TComplEx} & 0.560 & 47.0 & 61.0 & \multicolumn{1}{c|}{73.0}
        & 0.580 & 49.0 & 64.0 & \multicolumn{1}{c|}{76.0} & 0.731 & 67.3 & 76.2 & 84.5 \\
        \multicolumn{2}{c|}{TNTComplEx} & 0.560 & 46.0 & 61.0 & \multicolumn{1}{c|}{74.0}
        & 0.600 & 50.0 & 65.0 & \multicolumn{1}{c|}{78.0} & 0.718 & 65.4 & 74.9 & 85.6 \\
        \hline
        \multicolumn{2}{c|}{\method{}} & \textbf{0.610} & \textbf{50.9} & \textbf{67.7} & \multicolumn{1}{c|}{\textbf{79.0}}
        & \textbf{0.670} & \textbf{56.8} & \textbf{74.3} & \multicolumn{1}{c|}{\textbf{84.5}} & \textbf{0.778} & \textbf{69.7} & \textbf{84.4} & \textbf{90.3} \\
        \Xhline{2\arrayrulewidth}
    \end{tabular}}
    \caption{\method{} outperforms baselines in all three benchmarks, over all metrics. We used official implementation of DE-SimplE, and T(NT)ComplEx for Wikidata11k. Other results with [$\blacktriangledown$] are from \citet{ta-distmult}, results with [$\diamond$] are from \citet{de-simple}, and results on T(NT)ComplEx are from \citet{tcomplex}.}
    \label{tab:performance}
    \vspace{-4mm}
\end{table*}

%% file: 040experiment.tex
\subsection{Datasets}
We evaluate our proposed method on three benchmark datasets for TKG completion: ICEWS14, ICEWS05-15, and Wikidata11k, all suggested by \citet{ta-distmult}. ICEWS14 and ICEWS05-15 are subsets of ICEWS\footnote{https://dataverse.harvard.edu/dataverse/icews}, each containing socio-political events in 2014, and from 2005 to 2015 respectively. Wikidata11k is a subset of Wikidata\footnote{https://www.wikidata.org/wiki/Wikidata:Main\_Page}, posessing facts of various timestamps that span from A.D. 20 to 2020. All facts in Wikidata11k are annotated with additional temporal modifier, \textit{occurSince} or \textit{occurUntil}. For the sake of consistency and simplicity, we follow \citet{ta-distmult} to merge the modifiers into predicates rather than modeling them in separate dimension (e.g. (A, loves, B, since, 2020) transforms to (A, loves-since, B, 2020)). Detailed statistics of the three datasets are provided in Appendix B.

\subsection{Baselines}
We compare \method with representative static KG embeddings - TransE and DistMult, and state-of-the-art embedding -based baselines on temporal KGs, including ConT \cite{cont}, TTransE \cite{ttranse}, HyTE \cite{hyte}, TA \cite{ta-distmult}, DE-SimplE \cite{de-simple}, and T(NT)ComplEx \cite{tcomplex}.

\subsection{Experimental Setting}
For each dataset, we create $G_{KG}$ with only the triples in the train set. We add inverse edges to $G_{KG}$ for proper path-based inference on reciprocal relations. Also, we follow \citet{dpmpn} by adding self-loops to all entities in the graph, allowing the model to stay at the `answer node' if it reaches an optimal entity in $t < T$ steps.
To measure \method's performance in head entity prediction, we add reciprocal triples to valid and test sets too.
For all datasets, we find through empirical evaluation that setting the maximal path length $T = 3$ results in the best performance.
Following previous works, we fix the dimension of entity / relation / displacement embedding to 100.
Except for the embedding size, we search for the best set of hyperparameters using grid-based search, choosing the value with the best \textit{Hits@1} while all other hyperparameters are fixed.
We implement \method with PyTorch and DGL, and plan to make the code publicly available.
We provide further implementation details including hyperparameter search bounds and the best configuration in Appendix C.

\begin{figure*}[ht]
  \centering
  \includegraphics[width=\textwidth]{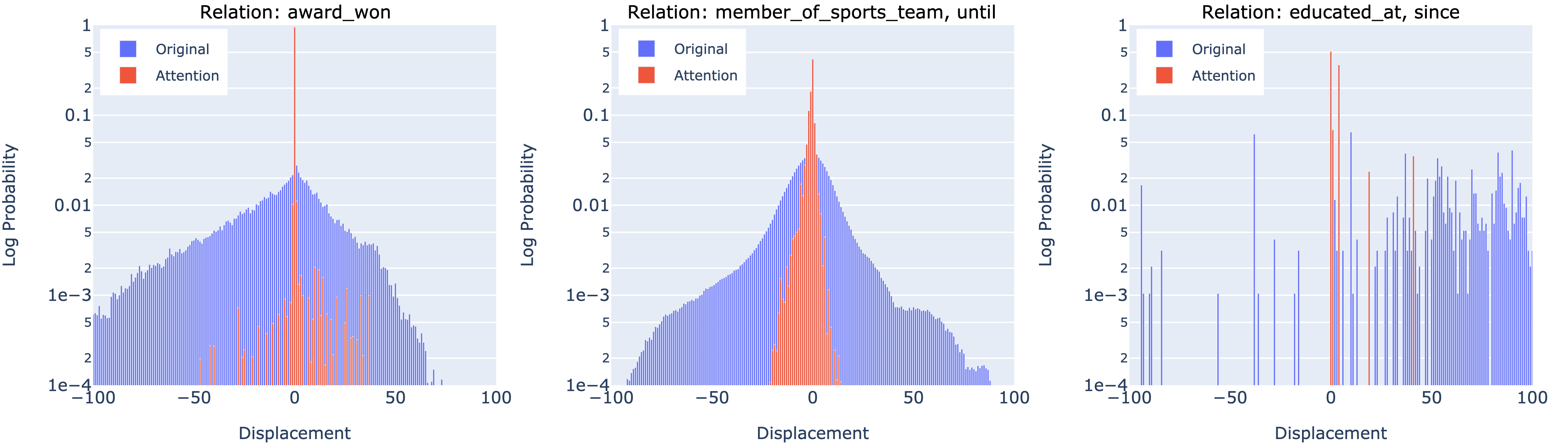}  
  \caption{Edge attention distribution over different temporal displacements. Note that y-axis is in log scale. For better visualization, we cut the range of temporal displacement to $[-100, 100]$ in all charts. {\method learns to effectively attend on events with specific temporal distance from the query time, depending on the relation type of the query.} }
  \label{fig:disp_distribution}
\end{figure*}

\subsection{Results}
Table \ref{tab:performance} shows the overall evaluation result of \method against baseline methods.
Along with Hits@1, 3, 10, we report MRR of the ground-truth entity, compared to the baseline methods.
As seen in the table, \method outperforms baseline models in all the benchmarks, improving up to 10\% relative performance consistently over all metrics.
We find through ablation study that the improvements are mainly contributed from resolving the two shortcomings of pre-existing TKG embeddings, which we indicate in earlier sections - absence of 1) neighborhood encoding scheme, 2) path-based inference with scalable subgraph sampling.  

\input{table/ablation}
\input{table/generalize}

\subsubsection{Model Variants and Ablation}
To further examine the effect of our proposed method in solving the aforementioned two problems, we conduct an ablation study as shown in Table~\ref{tab:ablation}.
First, we consider \method without temporal displacement encoding.
In both PGNN and SGNN, we do not consider the sign and magnitude of temporal displacement, and simply learn the embedding of each timestamp as is.
While computing the message $\vect{m}_{ij}$, the two GNNs simply add the time embedding $\gvect{\tau}_{t_{ij}}$ to $\vect{h}_i$ + $\gvect{\rho}_{ij}$.
No sign-specific weight is multiplied, and all edges are linear-transformed with a same weight matrix.
In this setting, \method's performance on ICEWS14 degrades about 30\% in Hits@1, and performs similar to TA-DistMult in Table~\ref{tab:performance}.
The result attests to the importance of temporal displacement for the neighborhood encoding in temporal KGs.

Next, to analyze the effect of subgraph sampling on overall performance, we resort to a new configuration of \method where no subgraph sampling is applied, and SGNN creates node feature $\vect{g}_i$ for all entities in $G_{KG}$.
Here, \method's performance slightly degrades about 1 percent in all metrics.
This implies the importance of subgraph sampling to prune query-irrevlant edges, helping \method to concentrate on the plausible substructure of the input graph.

Finally, we analyze the effect of PGNN by training \method with different numbers of PGNN layers.
We find that \method, trained with 1-layer PGNN, performs superior to the model without PGNN by absolute gain of 1\% in MRR.
However, adding up more layers in PGNN gives only a minor gain, or even aggravates the test set accuracy, mainly owing to early overfitting on the triples in train set.

\subsubsection{Generalizing to Unseen Timestamps}
We conduct an additional study that measures the performance of \method in generalizing to queries with unseen timestamps.
Following \citet{de-simple}, we modify ICEWS14 by including all triples except those on $5^{th}$, $15^{th}$, $25^{th}$ day of each month in the train set, and creating valid and test sets using only the excluded triples.
The performance of \method against the strongest baselines in this dataset are presented in Table~\ref{tab:generalize}.
In this setting, DE-SimplE and T(NT)ComplEx perform much more similar to each other than in Table~\ref{tab:performance}, while \method performs superior to all baselines.
DE-SimplE shows strength in generalizing over time, as it represents each entity as a continuous function over temporal dimension.
However, the model is weak when the range of timestamps is large and sparse, as shown for Wikidata in Table~\ref{tab:performance}.
Meanwhile, TComplEx and TNTComplEx show fair performance in Wikidata, but poorly infer for unseen timestamps, as they only learn independent embeddings of discrete timestamps.
On the contrary to these models, \method not only shows superior performance in all benchmarks but also is robust to unseen timestamps, by accounting for the temporal displacement, not the independent time tokens.

\subsection{Interpretation}
We provide a detailed analysis on the interpretability of \method in its relational reasoning process.

\subsubsection{Relation Type and Temporal Displacement}
Intuitively, the query relation type, and the temporal displacement between relevant events and the query are closely correlated.
For a query such as \textit{(PersonX, member\_of\_sports\_team, ?, $t_1$)}, events that happened 100 years before \textit{$t_1$} or 100 years after \textit{$t_1$} will highly likely be irrelevant.
On the contrary, for a query given as \textit{(NationX, wage\_war\_against, ?, $t_2$)}, one might have to consider those events far-off the time of interest.
To verify whether \method understands this implicit correlation, we analyze the attention distribution over edges with different temporal displacements, when \method is given input queries with a specific relation type.

The visualization of the distributions for three relation types are presented in Figure \ref{fig:disp_distribution}.
For all queries in the test set of WikiData11k with a specific relation type, we visualize the average attention value assigned to edges with each temporal displacement (red bars).
We compare this with the original distribution of temporal displacement, counted for all edges reachable in $T$ steps from the head entity $v_{query}$ (blue bars).
Remarkably, on the contrary with the original distribution of high variance over wide range of displacement, \method tends to focus most of the attention to edges with specific temporal displacement, depending on the relation type.

For queries with relation \textit{award\_won}, the attention distribution is extremely skewed, focusing over 90\% of the attention to events with displacement $= 0$ (i.e. events in the same year with the query).
Note that we have averaged the distribution for all queries with \textit{award\_won}, including both temporal modifiers \textit{occurSince} and \textit{occurUntil}.
The skewed distribution mainly results from the fact that the majority of the `award' entities in Wikidata11k are annual awards, such as \textit{Latin Grammy Award}, or \textit{Emmy Award}.
The annual property of the candidate entities naturally makes \method to focus on clues such as the nomination of awardees, or significant achievement of awardees in the year of interest.


\input{table/case_study}

Next, we test \method for queries with relation \textit{member\_of\_sports\_team-occurUntil}.
In this case, the attention is more evenly distributed than the former case, but slightly biased toward past events.
We find that this phenomenon is mainly due to the existence of temporally reciprocal edge in $G_{KG}$, which is a crucial key in solving the given query.
Here, \method sends more than half of the attention value (on average) to an event with relation \textit{member\_of\_sports\_team-occurSince}, that happened few years before the time of interest.
The inference follows our intuition to look for the last sports club where the player became member of, before the timestamp of the query.
The third case with relation \textit{educated\_at-occurSince} is the opposite of the second case.
Majority of the attention have been concentrated on events in 1-5 years after the query time, searching for the first event with relation \textit{educated\_at-occurUntil}, that happened after the time of interest.

As the analysis suggests, \method discovers important clues for each relation type, adequately accounting for the temporal displacement between the query and related events, while aligning with human intuition.


\subsubsection{Case Study}
We resort to a case study, to provide a detailed view on \method's attention-based traversal.
In this study, our model is given an input query \textit{(North\_Korea, threaten, ?, 2014/04/29)} in ICEWS14 where the correct answer is \textit{South\_Korea}.
For each propagation step, we list top-5 edges that received the highest attention value in the step.

The predominant edges and their associated attention values are shown in Table \ref{tab:case_study}.
In the first step, \method attends to various events pertinent to \textit{North\_Korea}, that mostly include negative predicates against other nations.
As seen in the table, the two plausible query-filling candidates are \textit{Japan}, and \textit{South\_Korea}. \textit{Japan} receives slightly more attention than \textit{South\_Korea}, as it is associated with more relevant facts such as ``\textit{North\_Korea} threatened \textit{Japan} at May 12th''.


In the second step, however, \method discovers additional relevant facts, that could be crucial in answering the given query.
As these facts have either \textit{Japan} or \textit{South \_Korea} as head entity, they could not be discovered in the first propagation step, which only propagates the attention from the query head \textit{North\_Korea}.
\method attends to the events \textit{(South\_Korea, threaten / criticize\_or\_denounce, North\_Korea)} that happened only a few days before our time of interest.
These facts imply the strained relationship between the two nations around the query time.
Also, \method finds that most of the edges that span from \textit{Japan} to \textit{North\_Korea} before/after few months the time of interest, tend to be positive events.
As a result, in the last step, \method propagates most of the node attention in \textit{North\_Korea} to the events associated with \textit{South\_Korea}.
Highest attention have been assigned to the relation \textit{make\_statement}.
Although the relation itself does not hold a negative meaning, in ICEWS14, \textit{make\_statement} is typically accompanied by \textit{threaten}, as entities do formally threaten other entities by making statements.

Through the case study, we find that \method leverages the propagation-based decoding as a tool to fix its traversal over wrongly-selected entities.
Although \textit{Japan} seemed like an optimal answer in the first step, it understands through the second step that the candidate was sub-optimal with respect to the query, propagating the attention assigned to \textit{Japan} back to \textit{North\_Korea}.
\method fixes its attention propagation in the last step, resulting in a completely different set of attended events compared to the first step.
Such an amendment would not have been possible with conventional approaches to path-based inference, which greedily select an optimal entity to traverse at each decoding step. 

%% file: table/ablation.tex
\begin{table}[t]
    \centering
    \small
    \begin{tabular}{ccccc}
        \Xhline{2\arrayrulewidth}
        \multicolumn{1}{c|}{Model} & \textit{MRR} & \textit{Hits@1} & \textit{Hits@3} & \textit{Hits@10} \\
        \hline\hline
        \multicolumn{1}{c|}{No Displacement} & 0.477 & 35.7 & 53.9 & 71.5 \\
        \multicolumn{1}{c|}{No Subgraph} & 0.598 & 49.3 & 66.8 & 78.5 \\
        \multicolumn{1}{c|}{No PGNN} & 0.59 & 49.9 & 66.5 & 78.4 \\
        \multicolumn{1}{c|}{2-layer PGNN} & 0.605 & 50.2 & 67.3 & \textbf{78.9} \\
        \hline
        \multicolumn{1}{c|}{\method} & \textbf{0.607} & \textbf{50.4} & \textbf{67.5} & 78.7 \\
        \Xhline{2\arrayrulewidth}
    \end{tabular}
    \caption{Ablation study results on ICEWS14, compared to the best configuration. Results annotated `\method' are the best performance of \method with temporal displacement encoding, 1-layer PGNN, and subgraph sampling.}
    \label{tab:ablation}
\end{table}

%% file: table/generalize.tex
\begin{table}[t]
    \centering
    \small
    \begin{tabular}{ccccc}
        \Xhline{2\arrayrulewidth}
        \multicolumn{1}{c|}{Model} & \textit{MRR} & \textit{Hits@1} & \textit{Hits@3} & \textit{Hits@10} \\
        \hline\hline
        \multicolumn{1}{c|}{DE-SimplE} & 0.434 & 33.3 & 49.2 & 62.4 \\
        \multicolumn{1}{c|}{TComplEx} & 0.443 & 34.8 & 49.2 & 62.5 \\
        \multicolumn{1}{c|}{TNTComplEx} & 0.444 & 34.6 & 49.4 & 63.5 \\
        \hline
        \multicolumn{1}{c|}{\method} & \textbf{0.483} & \textbf{37.2} & \textbf{54.6} & \textbf{69.0} \\
        \Xhline{2\arrayrulewidth}
    \end{tabular}
    \caption{Generalization performance over unseen timestamps in  ICEWS14. Accounting for relative displacement rather than independent timestamps, our model is the most robust to queries with unseen timestamps.}
    \label{tab:generalize}
\end{table}

%% file: table/case_study.tex
\begin{table*}[t]
    \centering
    \small
    \begin{tabular}{c|cc}
        \Xhline{2\arrayrulewidth}
        Query & \multicolumn{2}{c}{\textit{(North\_Korea, threaten, ?, 2014/04/29)}}\\
        \hline
        \multirow{5}{*}{1st Step}
        & \textit{(North\_Korea, \textcolor{red}{threaten}, Japan, 2014/05/12)} & 0.057 \\
        & \textit{(North\_Korea, \textcolor{red}{threaten}, Japan, 2014/12/18)} & 0.054 \\
        & \textit{(North\_Korea, make\_statement, South\_Korea, 2014/04/29)} & 0.044 \\
        & \textit{(North\_Korea, \textcolor{red}{make\_an\_appeal}, Japan, 2014/10/01)} & 0.041 \\
        & \textit{(North\_Korea, \textcolor{red}{threaten}, South\_Korea, 2014/08/01)} & 0.040 \\
        \hline
        \multirow{5}{*}{2nd Step}
        & \textit{(South\_Korea, \textcolor{red}{threaten}, North\_Korea, 2014/04/22)} & 0.15 \\
        & \textit{(Japan, \textcolor{blue}{release\_person}, North\_Korea, 2014/07/28)} & 0.078 \\
        & \textit{(South\_Korea, \textcolor{red}{criticize\_or\_denounce}, North\_Korea, 2014/04/28)} & 0.051 \\
        & \textit{(Japan, \textcolor{blue}{express\_intent\_to\_cooperate}, North\_Korea, 2014/02/28)} & 0.039 \\
        & \textit{(South\_Korea, make\_statement, North\_Korea, 2014/04/28)} & 0.037 \\
        \hline
        \multirow{5}{*}{3rd Step}
        & \textit{(North\_Korea, make\_statement, South\_Korea, 2014/04/29)} & 0.189 \\
        & \textit{(North\_Korea, \textcolor{red}{accuse}, South\_Korea, 2014/04/15)} & 0.121 \\
        & \textit{(North\_Korea, \textcolor{red}{criticized\_or\_denounced\_by}, South\_Korea, 2014/04/28)} & 0.052 \\
        & \textit{(North\_Korea, \textcolor{red}{deny\_responsibility}, South\_Korea, 2014/04/25)} & 0.032 \\
        & \textit{(South\_Korea, \textcolor{blue}{release\_person}, North\_Korea, 2014/04/14)} & 0.021 \\
        \hline
        Answer & \multicolumn{2}{c}{\textit{South\_Korea}}\\
        \Xhline{2\arrayrulewidth}
    \end{tabular}
    \caption{List of predominant edges for a case study. Numbers in the right are the corresponding edge attention assigned to each edge. Predicates in red color carry negative meaning, while predicates in blue color hold positive meaning. We find through the case study that the attention propagation allows \method to fix its misleading focus on sub-optimal entities.
    \vspace{-3mm}
    }
    \label{tab:case_study}
\end{table*} 

%% file: 050conclusion.tex
In this paper, we propose a novel approach to TKG completion named \method, which explores a query-relevant substructure of TKG with attention propagation.
Unlike other embedding-based models, the proposed method effectively gathers useful information from the existing KG, by accounting for the temporal displacement between the query and respective edges.
Quantitative results show that \method not only achieves state-of-the-art performance consistently over all three benchmarks, but also competently generalizes to queries with unseen timestamps.
Through extensive analysis, we also show that the propagated attention distribution well serves as an interpretable proxy of \method's reasoning process that aligns with human intuition.

%% file: 060appendix.tex
\appendix
\begin{figure*}[ht]
    \centering
    \includegraphics[height=9.5cm]{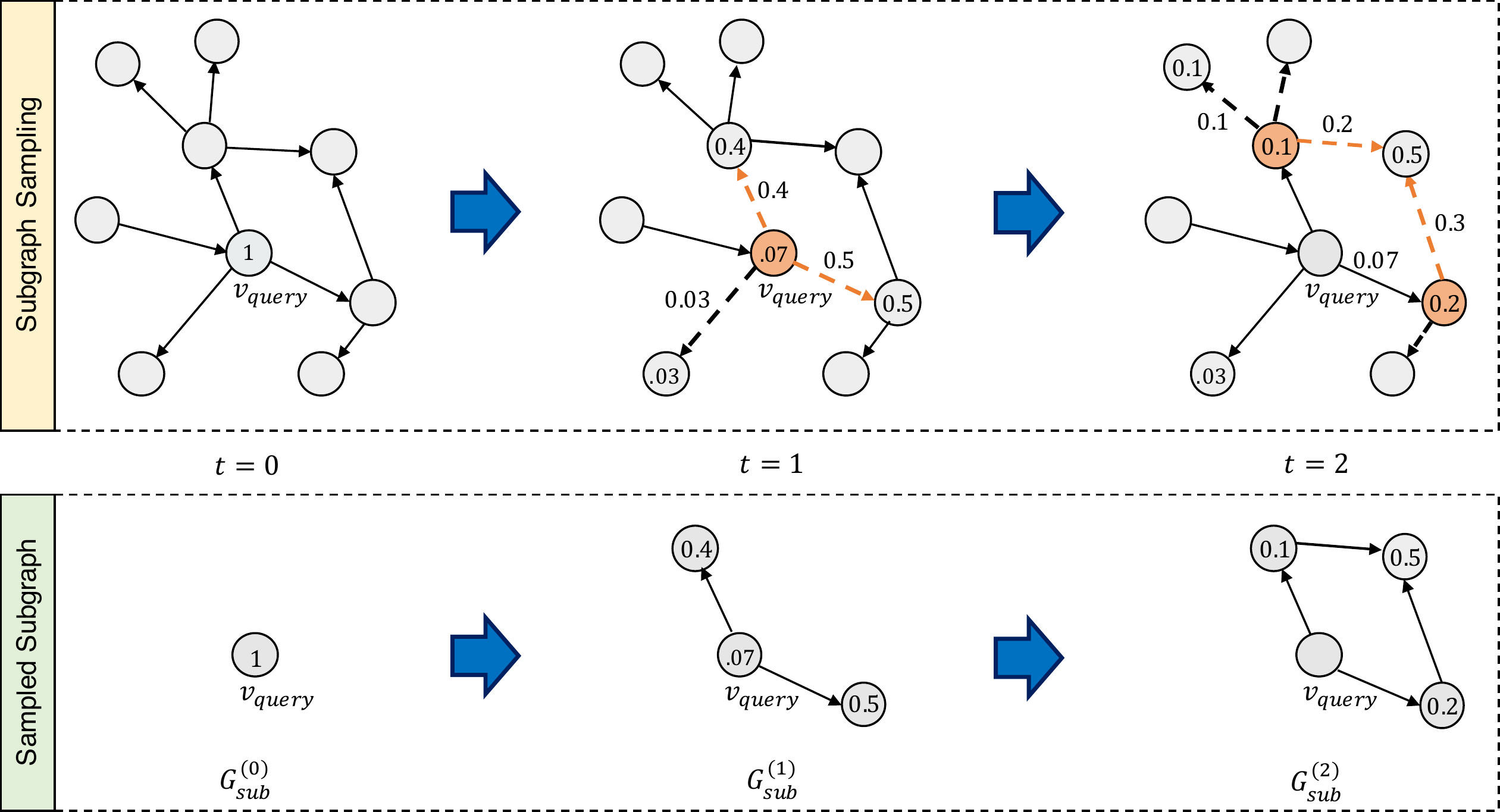}
    \caption{
    Example of subgraph sampling in \method.
    The graphs above represent the respective node / edge attention distribution at the initial state ($t=0$), after the first propagation step ($t=1$), and after the second propagation step ($t=2$).
    The graphs below show the sampled subgraph at each step $t$.
    $x$ $(=2)$ orange nodes are the core nodes retrieved at a step, and $y$ $(=3)$ dashed edges from each core node are candidate edges for the sampled subgraph.
    Among the candidate edges, $z$ $(=2)$ orange edges are newly added to the previous subgraph.}
    \label{fig:subgraph_example}
    \vspace{-3mm}
\end{figure*}

\lineskip=0pt
\section{\raggedright{A \quad Subgraph Sampling Example}} \label{sec:app_subgraph_sampling}
Figure~\ref{fig:subgraph_example} is an illustrative example for the subgraph sampling procedure in \method.
The hyperparameters for the example are as follows:
$x = 2$ (the maximum number of core nodes),
$y = 3$ (the maximum number of candidate edges, considered by each core node), and
$z = 2$ (the number of sampled edges
 added to the subgraph).

In the initial state, the only node associated with nonzero attention $a^{(0)}_i$ is the query head $v_{query}$.
Also, the initial subgraph $G^{(0)}_{sub}$ consists only of the node $v_{query}$.
After the first propagation step $t=1$, \method first finds top-$x$ core nodes (where $x=2$) w.r.t. nonzero  node attention scores $a^{(0)}_i$ of the previous step $t=0$.
Since the only node with nonzero attention value is $v_{query}$, it is retrieved as the core node.
Next, \method randomly samples at most $y = 3$ edges that originate from the core node (e.g. dashed edges with weights).
Among the sampled edges, it selects top-$z$ (where $z=2$) edges in the order of edge attention values $\widetilde{a}^{(1)}_{ij}$ at the current step; then, they are added to $G^{(0)}_{sub}$, resulting in the new subgraph $G^{(1)}_{sub}$.

After the second propagation step $t=2$, \method again finds $x$ core nodes that correspond to  highest attention values $a^{(1)}_i$ (e.g. nodes annotated with $0.1$ and $0.2$, respectively).
Then, $y$ outgoing edges for each core node are sampled; among $x \cdot y$ sampled edges, $z$ edges with highest edge attention values $\widetilde{a}^{(2)}_{ij}$ are added to $G^{(1)}_{sub}$, creating the new subgraph $G^{(2)}_{sub}$.

As seen in the figure, the incremental subgraph sampling scheme allows our model to iteratively expand the range of nodes and edges to attend, while guaranteeing that the critical nodes and edges in the previous steps are kept included in the latter subgraphs.

By flexibly adjusting the subgraph related hyperparameters $x, y$, and $z$, \method is readily calibrated between reducing computational complexity and optimizing the predictive performance. 
Intuitively, with more core nodes, more sampled edges, and more edges added to the subgraph, \method can better attend to the substructure of TKG that otherwise might have been discarded. 
Meanwhile, with small $x, y$, and $z$, \method can easily scale-up to large graphs by reducing the number of message-passing operations in SGNN. 

\newpage
\section{\raggedright{B \quad Dataset Statistics}} \label{sec:app_dataset_stat}
\input{table/dataset_stat}
\clearpage

\onecolumn
\section{\raggedright{C \quad Additional Implementation Detail}} \label{sec:app_impl_detail}
\input{table/impl_detail}

%% file: table/dataset_stat.tex
\begin{table}[htb]
    \centering
    \small
    \begin{tabular}{cccc}
        \Xhline{2\arrayrulewidth}
        \\[-1em]
        \multicolumn{1}{c|}{Dataset} & ICEWS14 & ICEWS05-15 & Wikidata11k \\
        \hline\hline
        \\[-1em]
        \multicolumn{1}{c|}{$|$\hspace{0.24mm}$V_{KG}$\hspace{0.24mm}$|$} & 7,128 & 10,488 & 11,134 \\
        \multicolumn{1}{c|}{$|R_{KG}|$} & 230 & 251 & 95 \\
        \multicolumn{1}{c|}{$|$\hspace{0.24mm}$T_{KG}$\hspace{0.24mm}$|$} & 365 & 4017 & 328  \\
        \multicolumn{1}{c|}{$|G_{KG}|$} & 90,730 & 479,329 & 150,079 \\
        \multicolumn{1}{c|}{\textit{Time Span}} & 2014 & 2005-2015 & 25-2020 \\
        \Xhline{2\arrayrulewidth}
    \end{tabular}
    \caption{Dataset Statistics for ICEWS14, ICEWS05-15, and Wikidata11k. The unit of the time span is year. }
    \label{tab:dataset_stat}
\end{table}

%% file: table/impl_detail.tex
\begin{table*}[ht]
    \centering
    \begin{tabular}{ ccc }
        \Xhline{2\arrayrulewidth} \\[-0.7em]
        \textbf{Computing Infrastructure} & \multicolumn{2}{c}{Tesla V100 GPU} \\[0.3em]
        \textbf{Search Strategy} & \multicolumn{2}{c}{Manual Tuning} \\[0.3em]
        \textbf{Best Validation \textit{Hits@1}} & \multicolumn{2}{c}{52.1 (ICEWS14), 56.8 (ICEWS05-15), 70.7 (Wikidata11k)} \\[0.3em]
        \Xhline{2\arrayrulewidth}
        &&\\
        \Xhline{2\arrayrulewidth} \\[-0.7em]
        \textbf{Hyperparameter} & \textbf{Search Bound} & \textbf{Experimental Setting} \\[0.3em]
        \hline \\[-0.8em]
        \textit{max path length $T$} & \textit{choice}[2, 3, 4] & 3 \\[0.3em]
        \textit{number $x$ of core nodes} & \textit{choice}[5, 10, 50, 100] & 100 (ICEWS14), 10 (ICEWS05-15), 50 (Wikidata11k) \\[0.3em]
        \textit{number $y$ of edges to sample} & \textit{choice}[10, 50, 100, 200, 500] & 500 (ICEWS14), 100 (ICEWS05-15), 200 (Wikidata11k) \\[0.3em]
        \textit{number $z$ of edges to add} & \textit{choice}[10, 50, 100, 200, 500] & 500 (ICEWS14), 100 (ICEWS05-15), 200 (Wikidata11k) \\[0.3em]
        \textit{number $K$ of attention heads} & \textit{choice}[3, 4, 5, 6] & 5 \\[0.3em]
        \textit{embedding dimension} & 100 & 100 \\[0.3em]
        \textit{number of epochs} & 10 & 10 \\[0.3em]
        \textit{batch size} & \textit{choice}[8, 16, 32] & 16 (ICEWS14), 8 (ICEWS05-15, Wikidata11k) \\[0.3em]
        \textit{optimizer} & \textit{Adam} & \textit{Adam} \\[0.3em]
        \textit{learning rate} & \textit{loguniform-float}[5e-2, 5e-5] & 5e-4 \\[0.3em]
        \textit{lr scheduler} & \textit{reduce\_on\_plateau} & \textit{reduce\_on\_plateau} \\[0.3em]
        \textit{lr reduction factor} & 0.1 & 0.1 \\[0.3em]
        \textit{gradient clip norm} & \textit{uniform-integer[1, 5]} & 3 (ICEWS14, ICEWS05-15), 5 (Wikidata11k) \\[0.3em]
        \Xhline{2\arrayrulewidth}
    \end{tabular}
    \caption{Additional implementation detail of \method. We report the hyperparameter search bounds and the used configurations, along with the experimental environments.}
    \label{tab:impl_detail}
\end{table*}